\documentclass[default,iicol]{sn-jnl}% Default with double column layout
\usepackage{lipsum}

\newcommand\blfootnote[1]{%
  \begingroup
  \renewcommand\thefootnote{}\footnote{#1}%
  \addtocounter{footnote}{-1}%
  \endgroup
}

\usepackage{natbib}
\usepackage{lscape}
\usepackage{longtable}
\usepackage{comment}
\usepackage{algpseudocode}
\usepackage{algorithm}
\usepackage{soul}
\usepackage{caption}

\jyear{2022}%

\theoremstyle{thmstyleone}%
\theoremstyle{thmstyletwo}%

\theoremstyle{thmstylethree}%

\begin{document}

\title[Learning Semantic Text Similarity to rank Hypernyms of Financial Terms]{Learning Semantic Text Similarity to rank Hypernyms of Financial Terms}


\author[1]{\fnm{Sohom} \sur{Ghosh}\textsuperscript{*,}}\email{sohom1ghosh@gmail.com}
\author[2]{\fnm{Ankush} \sur{Chopra}\textsuperscript{$\dagger,$}}\email{ankush01729@gmail.com} \author[3]{\fnm{Sudip Kumar} \sur{Naskar}}\email{sudip.naskar@gmail.com}

\affil[1]{\orgname{Fidelity Investments}, \orgaddress{ \city{Bengaluru}, \state{Karnataka}, \country{India}}, ORCiD: 0000-0002-4113-0958}
\affil[2]{\orgname{Tredence Analytics}, \orgaddress{ \city{Bengaluru}, \state{Karnataka}, \country{India}}, ORCiD: 0000-0002-9970-8038}
\affil[1,3]{\orgname{Jadavpur University}, \orgaddress{ \city{Kolkata}, \state{West Bengal}, \country{India}}, ORCiD: 0000-0003-1588-4665}

% \author*[1,2]{\fnm{First} \sur{Author}}\email{iauthor@gmail.com}



% \affil*[1]{\orgdiv{Department}, \orgname{Organization}, \orgaddress{\street{Street}, \city{City}, \postcode{100190}, \state{State}, \country{Country}}}

% \affil[2]{\orgdiv{Department}, \orgname{Organization}, \orgaddress{\street{Street}, \city{City}, \postcode{10587}, \state{State}, \country{Country}}}

% \affil[3]{\orgdiv{Department}, \orgname{Organization}, \orgaddress{\street{Street}, \city{City}, \postcode{610101}, \state{State}, \country{Country}}}


\abstract{Over the years, there has been a paradigm shift in how users access financial services. With the advancement of digitalization more users have been preferring the online mode of performing financial activities. This has led to the generation of a huge volume of financial content. Most investors prefer to go through these contents before making decisions. Every industry has terms that are specific to the domain it operates in. Banking and Financial Services are not an exception to this. In order to fully comprehend these contents, one needs to have a thorough understanding of the financial terms. Getting a basic idea about a term becomes easy when it is explained with the help of the broad category to which it belongs. This broad category is referred to as \textit{hypernym}. For example, ``bond'' is a hypernym of the financial term ``alternative debenture''.
In this paper, we propose a system capable of extracting and ranking hypernyms for a given financial term. The system has been trained with financial text corpora obtained from various sources like DBpedia \cite{dbpedia}, Investopedia, Financial Industry Business Ontology (FIBO), prospectus and so on. Embeddings of these terms have been extracted using FinBERT \cite{araci2019finbert}, FinISH \cite{akl-etal-2021-yseop} and fine-tuned using SentenceBERT \cite{sbert}. A novel approach has been used to augment the training set with negative samples. It uses the hierarchy present in FIBO. Finally, we benchmark the system performance with that of the existing ones. We establish that it performs better than the existing ones and is also scalable. Our code base will be made available at: \url{https://github.com/sohomghosh/FinSim_Financial_Hypernym_detection} after acceptance of this manuscript.}




\maketitle
\blfootnote{\textsuperscript{$\dagger$}This work was done when Ankush was previously associated with Fidelity Investments, India} %%%comment out remove later SOHOM
\blfootnote{This paper is an extension of the solution \cite{chopra-ghosh-2021-term} presented by our team LIPI at FinSim-3 \cite{kang-etal-2021-finsim} (FinNLP-2021 - workshop of IJCAI-2021)}. %%%comment out remove later, OUR TEAM LIPI SOHOM
\blfootnote{*Corresponding Author} %%%comment out remove later SOHOM
\blfootnote{This pre-print has not undergone peer review (when applicable) or any post-submission improvements or corrections. The Version of Record of this article is published in Springer Nature Computer Science, and is available online at \url{https://doi.org/10.1007/s42979-023-02134-z}}

\clearpage
\section{Introduction}
\label{sec:introduction}
\begin{figure}[ht]
  \centering
  \includegraphics[width=6cm]{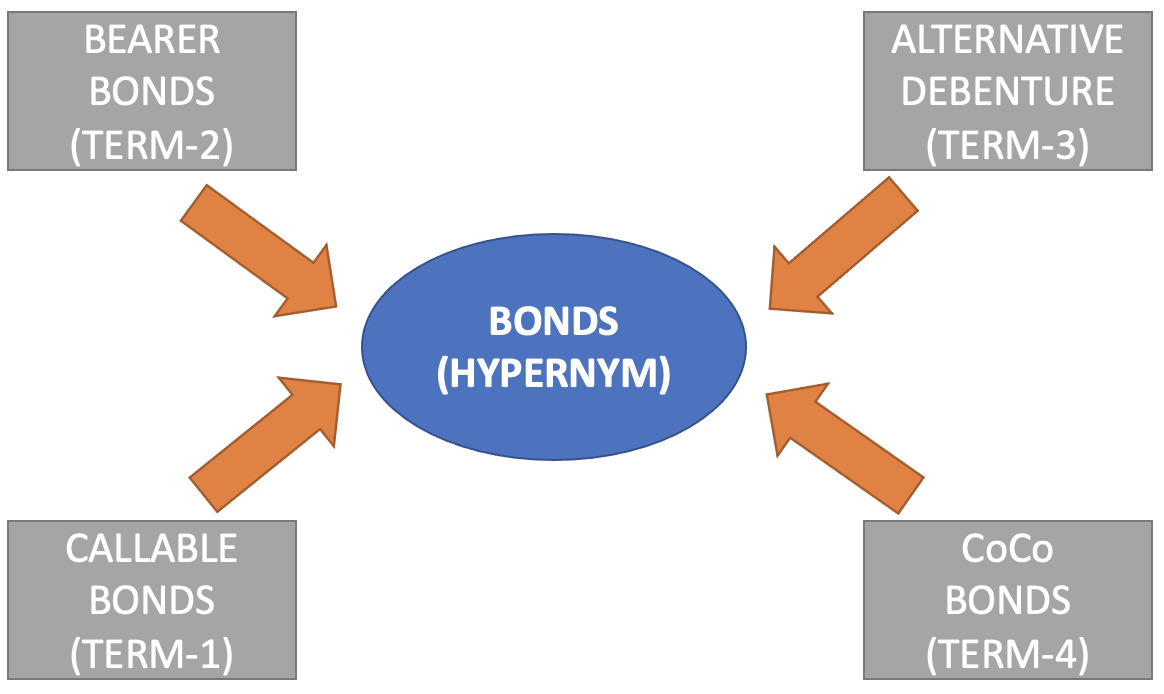}
  \caption{Terms to Hypernym relation}
  \label{fig:termhyp}
\end{figure}

Investors read online content (like financial reports of organizations, news) to make decisions. These contents often contain jargon unknown to the readers. The readability of these contents can be improved significantly by presenting readers with hypernyms (i.e. board categories) corresponding to any jargon. A jargon being a subset holds an ``IS A'' relationship with its hypernym. For example, ``alternative debenture'' (unknown financial term/jargon) is a kind of ``bond'' (hypernym). The same holds true for terms like ``Bearer Bonds'', ``Callable Bonds'' and ``CoCo Bonds''. This is shown in Figure \ref{fig:termhyp}. The Natural Language Processing (NLP) community has been working on methods to automatically discover hypernyms for more than a decade. Recently with the advent of shared tasks like FinSim \cite{maarouf-etal-2020-finsim} extracting hypernyms specific to the financial domain has caught the attention of this community. Inspired by the advances and contributions made by the participants in FinSim-1 \cite{maarouf-etal-2020-finsim} and FinSim-2 \cite{all-2021-finsim}, we participated in the third edition of FinSim \cite{kang-etal-2021-finsim}.
It comprised of matching financial terms to their hypernyms. Compared to the previous two editions, the third edition consisted of larger and more diverse topics related to finance. In this paper, we present an extension of the solutions our team LIPI developed while participating in FinSim-3 as well as the enhancements we carried out later. %% SOHOM OUR TEAM LIPI

\subsection*{Research Questions}
The research questions we try to answer in this study are as follows.
\begin{itemize}
    \item \textbf{RQ1:} How have the datasets and solution architectures of the FinSim challenges evolved over the years?
    \item \textbf{RQ2:} How to develop a system for ranking a set of hypernyms for a given financial term?
    \item \textbf{RQ3:} Does using domain specific embeddings  improve model performance?
    \item \textbf{RQ4:} What is the impact of augmenting/adding data from other sources?
\end{itemize}

\subsection*{Our Contributions}
Our contributions in the work contained in this article are as follows.
\begin{itemize}
    \item We review and summarize various approaches used by participants of all three editions of FinSim \cite{maarouf-etal-2020-finsim, all-2021-finsim, kang-etal-2021-finsim}. We further collate the performances of such approaches in Table \ref{tab:related-works-finsim}.
    \item We explore various external financial data sources to supplement the training set.
    \item We propose a novel way of augmenting the training set for incorporating hierarchies that are present in the set of hypernyms.
    \item We develop a system capable of ranking a set of hypernyms for a given financial term.
    
\end{itemize}
\subsection*{Reproducibility} The data set used in this paper can be obtained from here\footnote{\url{https://sites.google.com/nlg.csie.ntu.edu.tw/finnlp2021/shared-task-finsim}}. The metadata is presented in the paper \cite{kang-etal-2021-finsim}.  Our code base will be made available\footnote{\url{https://github.com/sohomghosh/FinSim_Financial_Hypernym_detection}} after acceptance of the paper. %[ADD CODE TO GIT] %%SOHOM

\subsection*{Structure of the paper}
This paper is organized as follows. Section \ref{sec:introduction} introduces readers to our motivation. Section \ref{sec:literature} briefly narrates the previous works on this task. We formally define the problem statement in Section \ref{sec:problem} and discuss the dataset used for this work in Section \ref{sec:dataset}. Next, we describe our methodology, experiments and results in Section \ref{sec:meth}, \ref{sec:experimentation} and \ref{sec:results}  respectively. Section \ref{sec:conclusion} concludes the paper and section \ref{sec:future} provides avenues for future work.

\section{Research Landscape}
\label{sec:literature}
In this section, we discuss the previous works in three phases. Firstly, we explore how the problem of hypernym identification have been solved in the field of computational linguistics in general. Following this, we elaborate its applications specific to the Financial Domain. Finally, we state how our work differs from the existing work in the literature.

\subsection*{Hypernym Identification in NLP Literature}
The task of Hypernym detection started gaining the interest of the NLP community in early 1990. During this time Hearst et al. \cite{hearst-1992-automatic} did the pioneering work of automatically extracting hypernyms using lexico-syntactic patterns like ``such as'' followed and preceded by Noun Phrase and so on. Another pattern-based approach had been applied by Snow et al. \cite{snow2005learning}. They narrated how they extracted ``dependency paths''from parse trees of sentences containing hypernyms and hyponyms using WordNet \cite{miller1998wordnet}. They additionally used coordinate terms i.e. terms having at least one common parent to enhance the process of hypernym identification. Sang  \cite{tjong-kim-sang-2007-extracting} assumed that the web contained much more data than any of the text corpora and developed a simple pattern-based method to extract hypernyms from the web. Furthermore, Sang et al.\cite{tjong-kim-sang-hofmann-2009-lexical} compared two major approaches of hypernym extraction which are based on lexical (dictionary-based) and dependency patterns. Ritter et al. \cite{ritter2009anyway} described how they used lexical based patterns and Hidden Markov Models to identify hypernyms of noun phrases. 

Distributional representation of a text corpus refers to the process of representing texts as vectors. Papers \cite{yamada-etal-2009-hypernym, espinosa-anke-etal-2016-supervised, yamane-etal-2016-distributional} narrate various distributional frameworks for mining hypernyms from sources like Wikipedia \cite{wiki}. Yamada et al. \cite{yamada-etal-2009-hypernym} proposed two methods for calculating similarities in distributional representation for extracting relations. One of them used the raw verb-noun dependency while the other one used clusters from the dependency. Furthermore, they evaluated this on Japanese Web pages. Espinosa-Anke et al. %\textcolor{blue}{\cite{camacho-collados-etal-2018-semeval}} 
Camacho-Collados et al. \cite{camacho-collados-etal-2018-semeval} proposed TAXOEMBED2 which was a distributional representation framework created from sense embeddings. Furthermore, it learned semantics from any given domain. Weeds et al. \cite{weeds-etal-2014-learning} studied relationships between two words given their distributional vectors using a SVM. Roller et al. \cite{roller-etal-2014-inclusive} validated ``Distributional Inclusion Hypothesis'' for extracting hypernyms and concluded that it works only when applied on a set of dimensions that are related. Thus, they proposed the ``Selective Distributional Inclusion Hypothesis''. Yamane et al. proposed a model \cite{yamane-etal-2016-distributional} that firstly created clusters for hypernym creation. After that, it adjusted the number of clusters and used negative samples that are non-hypernym instances for learning.

In another paper, Roller et al. \cite{roller-etal-2018-hearst} compared both the approaches mentioned above: pattern based and distribution-based approaches. They established that pattern-based approaches perform better especially when dealing with context-based
constraints. Shwartz et al. \cite{shwartz-etal-2016-improving} ensembled these two approaches using Long Short Term Memory Network (LSTM) \cite{lstm}. Similarly, Held et al. \cite{held-habash-2019-effectiveness} created a hybrid model by combining these approaches. They used the nearest neighbour algorithm in their distribution-based approach. In the paper \cite{cho-etal-2020-leveraging}, Cho et al. proposed a model hypo2path which was an encoder-decoder model trained using paths from WordNet.

Caraballo \cite{caraballo1999automatic} presented an automatic method of building a hierarchy of nouns and their hypernyms using WordNet \cite{miller1998wordnet}. Bottom-up clustering had been used to create the hierarchy and hypernyms had been assigned after creating a binary tree. Shinzato et al. \cite{shinzato-torisawa-2004-acquiring} proposed a novel method of extracting hypernyms from web pages using structures of the HTML pages and other statistical features. They further extend this study specifically for Japanese HTML documents \cite{shinzato-torisawa-2004-extracting}. Dias et al. \cite{dias-2008-mapping} used directed graphs (weighted as well as unweighted) and combined it with TextRank \cite{mihalcea-tarau-2004-textrank} algorithm for inferring various relations between nouns present in the Web. Navigli et al. \cite{navigli-velardi-2010-learning} introduced a novel concept of Word-Class Lattices which were learned from definitions present in Wikipedia. They further released a Java-based tool \cite{faralli-navigli-2013-java} to extract hypernyms of a term and its' definitions. Boella et al. \cite{boella-di-caro-2013-extracting} firstly identified sentences which are definitional. Subsequently, they extracted syntactic features from these sentences using the parser they developed. They later fed these features to a Support Vector Machine \cite{svm} based classifier to identify hypernyms.

The problem of Hypernym detection has been explored in languages other than English as well. For instance, Lefever et al. \cite{lefever-etal-2014-evaluation} and Yildirim et al. \cite{yildirim-yildiz-2012-automatic} extended it to Dutch and Turkish corpora respectively. Lefever et al. \cite{lefever-etal-2014-evaluation} evaluated three approaches - morphology-based, dictionary-based and distribution based. Out of these, the first one performed the best on dredging and financial domain specific data. Yildrim et al. used dictionary-based syntactic patterns as well as semantic similarity. Grycner et al. \cite{grycner-etal-2015-relly} proposed ``RELLY'' a method to construct a graph comprising hypernyms of relational phrases. Gupta et al. \cite{gupta-etal-2017-cikm-taxonomy} developed a semi-supervised approach to extract sub-sequences of hypernyms using a list of seed terms as input. Fu et al. \cite{fu-etal-2013-exploiting} worked on hypernym discovery on Chinese encyclopedias. They manually annotated a dataset consisting of Chinese terms. After extracting hypernyms, they developed a statistical system to rank them.
Recently, the use of Deep Learning Models in Computational Linguistics has gathered the interest of the NLP community. Tan et al. \cite{tan-etal-2020-from-syntactic} used bi-directional Recurrent Neural Networks to extract hypernyms from definitions using Parts of Speech of constituent words. They validated this model's performance on Wikipedia as well as Stack-Overflow datasets. %\textcolor{blue}{Liang et al.} 
Liang et al. \cite{liang-etal-2017-aaai-on} studied if the property of transitivity holds in lexical taxonomies which were built automatically. They developed a supervised approach to do so. Furthermore, they used transitivity to extract new hypernym-hyponym relations.

\subsection*{SemEval Shared Tasks on Hypernym Detection}
Problems relating to hypernym detection were provided in several editions of SemEval \cite{bordea-etal-2015-semeval, bordea-etal-2016-semeval, augenstein-etal-2017-semeval, camacho-collados-etal-2018-semeval}. 

\textbf{SemEval-2015 Task 17}: ``Taxonomy Extraction Evaluation (TExEval)" \cite{bordea-etal-2015-semeval} dealt with extraction of hypernym-hyponym relations  from texts and taxonomy construction for four different domains namely: chemicals, equipment, foods and science. Grefenstette \cite{grefenstette-2015-inriasac} developed the best performing model using simple structure-based features like whether a term is present in a sentence and document, term and document frequencies and presence of sub-sequences.

\textbf{SemEval-2016 Task 13}: ``Taxonomy Extraction Evaluation (TExEval-2)" \cite{bordea-etal-2016-semeval} was the multilingual edition of TExEval \cite{bordea-etal-2015-semeval}. It comprised corpora from several domains like environment, food and science. Different languages included English, Dutch, Italian and French. Team Taxi \cite{panchenko-etal-2016-taxi} won both the shared tasks. They used Hearest pattern and sub-string based features.

\textbf{SemEval 2017 Task 10}: ``ScienceIE - Extracting Keyphrases and Relations from Scientific Publications'' \cite{augenstein-etal-2017-semeval} dealt with extraction of important phrases (like Process, Task and Material) and relations (like hypernyms / synonyms). It was restricted to the scientific domain. Team MIT \cite{lee-etal-2017-mit} achieved the first rank by creating a system using a convolutional neural network. This system used an embedding comprising relative positions, type of entity and parts of speech as input.

\textbf{SemEval-2018 Task 9}: ``Hypernym Discovery'' was introduced \cite{camacho-collados-etal-2018-semeval} in the year 2018. This shared task was about extracting hypernyms from corpora in three languages (English, Spanish and Italian) and two domains within English (Medical and Music). The best performing model was presented by Team CRIM \cite{bernier-colborne-barriere-2018-crim}. This model was an ensemble of word embedding based supervised approach with a pattern based unsupervised approach.

Dash et al. \cite{DBLP:conf/aaai/DashCGMF20} introduced a new neural network-based architecture, Strict Partial Order Networks (SPON) to detect hypernyms. They benchmarked it using SemEval 2018 general and domain specific hypernym discovery tasks. Very recently Bai et al. \cite{bai-etal-2021-hypernym} proposed the use of sequential recurrent mapping models to preserve the hierarchy between terms and their hypernyms. They also performed an extensive evaluation on SemEval-2018 Task 9 datasets.

\subsection*{FinSim Shared Tasks - Hypernym Detection in Financial Texts}

\begin{table*}
\centering
\caption{Background. \#Pps is number of Prospectus. \#L, \#T, Acc. and MR. denote number of Labels, Teams, Best Accuracy and Mean Rank respectively.}
\label{tab:background-table}
%\begin{tabular}{|l|l|l|l|l|l|l|l|l|l|}
\begin{tabular}{lllrrrrrrr}
%\hline
\toprule
\textbf{Year} & \textbf{Edition} & \textbf{Conference} & \textbf{\#Pps} & \textbf{\#Train} & \textbf{\#Test} & \textbf{\#L} & \textbf{\#T} & \textbf{Acc.} & \textbf{MR.} \\ %\hline
 \midrule
2020          & FinSim-1 \cite{maarouf-etal-2020-finsim}      & IJCAI-PRICAI   & 156                   & 100              & 99              & 8                 & 6                & 0.858              & 1.21              \\ %\hline
2021          & FinSim-2 \cite{all-2021-finsim}      & ACM-WWW            & 203                   & 614              & 211             & 10                & 7                & 0.906              & 1.189             \\ %\hline
2021          & FinSim-3 \cite{kang-etal-2021-finsim}      & IJCAI            & 211                   & 1050              & 326             & 17                & 5                & 0.941              & 1.113             \\ %\hline
\bottomrule
\end{tabular}
\end{table*}

As mentioned earlier, the third edition of FinSim challenge \cite{kang-etal-2021-finsim} is the most recent one. Details relating to all editions of FinSim is mentioned in Table \ref{tab:background-table}. These shared tasks have been organized by  Fortia Financial Solutions\footnote{\url{https://www.fortia.fr/}}. Teams IITK \cite{keswani-etal-2020-iitk}, PolyU-CBS \cite{pplyu-2021-finsim} and MXX \cite{kroher-etal-2021-mxx} won the first, second and third editions of FinSim respectively. We shall narrate more details relating to the dataset of FinSim-3 in the next section \ref{sec:dataset}. Team MXX \cite{kroher-etal-2021-mxx} used a LSTM \cite{lstm} based approach over word2vec \cite{word2vec} embeddings to win the FinSim-3 challenge (Accuracy = 1.113, Mean Rank = 0.941). The evaluation metrics and the other aspects of the problem statement remained the same for all three editions. We organize the system descriptions of the participating teams and present them in Table \ref{tab:related-works-finsim}. The winning entries have been highlighted in bold. Studying this table thoroughly, we observe that the Word2Vec approach remained the same for all of them. Only one of these teams MXX \cite{kroher-etal-2021-mxx} augmented the given dataset with external data. Similarly, only one of the winning team PolyU-CBS used syntactic based features like Jaccard similarity. Logistic Regression emerged out to be the most preferred classifier. Moreover, it is interesting to note that every successive year performances of the submitted models improved significantly. Since only three teams (\cite{portisch2021finmatcher}, \cite{jsi-2021-finsim} and \cite{feng-wei-2021-exploiting}) used Knowledge Graphs, we conclude it is yet to become popular. Some of the BERT based models like FinBERT \cite{araci2019finbert}, Sentence BERT \cite{sbert} and RoBERTa \cite{Liu2019RoBERTaAR} were also explored by most participants.

In recent times, Loukas \cite{loukas2021edgarcorpus} released the EDGAR-CORPUS comprising annual reports of listed US organizations from the year 1993 to 2020. They created word2vec \cite{word2vec} embeddings based on this corpus and evaluated it on the FinSim-3 dataset. They achieved an accuracy of 0.879 and a mean average rank of 1.21 using stratified 10-fold cross-validation.

\subsection*{Difference with Prior Works}
Our work is novel in terms of the approach we used to create negative samples from the existing dataset using the hierarchy present within the hypernyms. Unlike most others, we did not train a classifier to solve the problem of detecting hypernyms. On the other hand, we detect hypernyms by performing semantic search over fine-tuned embeddings. This makes the approach generic and robust to adding more hypernyms to the existing set.

\section{Problem Statement}
\label{sec:problem}
In this section, we shall narrate the problem statement and discuss the evaluation metrics.

\subsection*{Problem Definition}
Given a set of n financial terms (t\textsubscript{1}, t\textsubscript{2}, t\textsubscript{3}, ... t\textsubscript{n}) and their corresponding hypernyms/labels (l\textsubscript{1}, l\textsubscript{2}, l\textsubscript{3}, ... l\textsubscript{n}) where $l\textsubscript{i} \epsilon$ \{Equity Index, Regulatory Agency, Credit Index, Central Securities Depository, Debt pricing and yields, Bonds, Swap, Stock Corporation, Option, Funds, Future, Credit Events, MMIs, Stocks, Parametric schedules, Forward, Securities restrictions\}. Our task is to develop a system capable of ranking all these hypernyms  in order of decreasing semantic similarity for any unknown financial term.

\subsection*{Evaluation Metrics}
The evaluation metrics used here are as follows:\\
$Accuracy = \frac{1}{n}*\sum_{i=1}^{n}I(y_i=\hat{y_i}[1])$,\\%\textcolor{blue}{,}\\
$Mean Rank = \frac{1}{n}*\sum_{i=1}^{n}(\hat{y_i}.index(y_i))$,\\%\textcolor{blue}{,}\\
 where $\hat{y_i}$ is the ranked list (with the index starting from 1) of predicted labels corresponding to the expected label $y_i$. I is an identity matrix. Interestingly, the organizers considered only the first three elements of the ranked list for evaluation. If any label was not present within these three elements, it was assigned rank 4.

% \usepackage{lscape}
% \usepackage{longtable}
% Note: It may be necessary to compile the document several times to get a multi-page table to line up properly
\clearpage
\onecolumn
\begin{landscape}
%\sidewaystablefn%
\begin{longtable}[c]{|l|l|r|r|l|l|l|l|l|}
%%%\caption{CAPTION INSERT}

\caption{Related Works - FinSim. USE: Universal Sentence Encoder, RF: Random Forest, LR: Logistic Regression, LSTM: Long Short Term Memory; NB: Naive Bayes, NN: Neural Networks, DA: Deep Attention, KG: Knowledge Graphs, SVM: Support Vector Machine, Inv: Investopedia, Ext: External}
\label{tab:related-works-finsim}\\
\hline
                  &                          & \multicolumn{1}{l|}{}              &             &                                                                                                                \multicolumn{5}{c|}{\textbf{Approach of best performing model}}                                                                                                                                \\ \hline
\endfirsthead
\multicolumn{9}{c}%
{{\bfseries Table \thetable\ continued from previous page}} \\
\hline
                  &                          & \multicolumn{1}{l|}{}              & \multicolumn{1}{l|}{}            &                                                                                                               & \multicolumn{4}{c|}{\textbf{Approach of best performing model}}                                                                                                                                \\ \hline
\endhead
\textbf{Task}     & \textbf{Team}            & \multicolumn{1}{l|}{\textbf{Acc.}} & \multicolumn{1}{l|}{\textbf{MR}} & \textbf{Syntatic Features}                                                                                    & \textbf{Classifier} & \textbf{Embeddings}                                                    & \textbf{KG} & \textbf{\begin{tabular}[c]{@{}l@{}}Ext. Data\end{tabular}}                                                     \\ \hline
FinSim-1          & Anuj \cite{saini-2020-anuj-investopedia}                     & 0.858                              & 1.42                             & \begin{tabular}[c]{@{}l@{}}Character count,\\ Word Count etc.\end{tabular}                                    & SVM                 &                                                                        &             & Inv                                                                      \\ \hline
FinSim-1          & ProsperaMnet \cite{berend-etal-2020-prosperamnet}             & 0.777                              & 1.34                             &                                                                                                               &                     & Sparse embeddings                                                      &             &                                                                                   \\ \hline
FinSim-1          & FINSIM20 \cite{anand-etal-2020-finsim20}                & 0.787                              & 1.43                             &                                                                                                               &                     & USE \cite{cer2018universal}                                                                    &             &                                                                                   \\ \hline
\textbf{FinSim-1} & \textbf{IIT-K} \cite{keswani-etal-2020-iitk}          & \textbf{0.858}                     & \textbf{1.21}                    &                                                                                                               & NB         & Word2Vec, BERT                                                &             &                                                                                   \\ \hline
FinSim-2          & AIAI \cite{aiaifinsim2} & 0.877                              & 1.278                            &                                                                                                               & DA                  & \begin{tabular}[c]{@{}l@{}}Word2Vec \cite{word2vec}\end{tabular} &             &                                                                                   \\ \hline
FinSim-2          & FinMatcher \cite{portisch2021finmatcher}              & 0.811                              & 1.415                            & Word overlap                                                                                                  & NN                  &                                                                        & RDF2vec     & \begin{tabular}[c]{@{}l@{}}WordNet,\\Wikidata,\\ WebIsALOD\end{tabular}            \\ \hline
FinSim-2          & GOAT \cite{goat-2021-finsim}                    & 0.896                              &1.193                             &                                                                                                               & LR                  & \begin{tabular}[c]{@{}l@{}}FinBERT \cite{araci2019finbert}\end{tabular}  &             & Inv                                                                      \\ \hline
FinSim-2          & L3i-LBPAM  \cite{l3i-2021-finsim}              &  0.858                             & 1.325                            &                                                                                                               &                     & SentenceBERT \cite{sbert}                                                          &             &                                                                                   \\ \hline
FinSim-2          & TCSWTIM2021 \cite{tcs2021finsim}             &  0.858                             & 1.274                            & Sentence extraction                                                                                           &               & TF-IDF, BERT \cite{devlin-etal-2019-bert}                                                                  &             &                                                                                   \\ \hline
FinSim-2          & JSI  \cite{jsi-2021-finsim}                    &  0.811                             & 1.316                            &                                                                                                               & RF                  & Word2vec                                                               & FIBO        &                                                                                   \\ \hline
\textbf{FinSim-2} & \textbf{PolyU-CBS} \cite{pplyu-2021-finsim}      &  \textbf{0.906}                    & \textbf{1.189}                   & \begin{tabular}[c]{@{}l@{}}Jaccard similarity,\\  if hypernym in term\end{tabular}                   & LR         & Word2Vec, BERT                                                &             &                                                                                   \\ \hline
FinSim-3          & DICoE  \cite{loukas-etal-2021-dicoe}                  &  0.904                             & 1.162                            & \begin{tabular}[c]{@{}l@{}}Levenshtein distance,\\  Upper case to lower \\ case characters ratio\end{tabular} & LR                  & Word2vec                                                               &             & \begin{tabular}[c]{@{}l@{}}Inv,\\ FIBO\end{tabular}                                                                \\ \hline
FinSim-3          & MiniTrue  \cite{feng-wei-2021-exploiting}               & 0.865                              &   1.315                          &                                                                                                               &                     & BERT, FinBERT                                                          & RotatE      &                                                                                   \\ \hline
FinSim-3          & \begin{tabular}[c]{@{}l@{}}Lipi  \cite{chopra-ghosh-2021-term} \\ \textit{(Our old model)} \end{tabular}                 &   0.917                            & 1.156                            & Acronym expansion                                                                                             &                     & SentenceFinBERT                                                        &             & \begin{tabular}[c]{@{}l@{}}DBPedia,\\Inv,\\ FIBO\end{tabular}             \\ \hline
FinSim-3          & Yseop \cite{akl-etal-2021-yseop}                   &   0.917                            & 1.141                            &                                                                                                               & LR                  & \begin{tabular}[c]{@{}l@{}}FastText, \\ SentenceRoBERTa\end{tabular}   &             & FIBO                                                                              \\ \hline
\textbf{FinSim-3} & \textbf{MXX} \cite{kroher-etal-2021-mxx}            &  \textbf{0.941}                    & \textbf{1.113}                   &                                                                                                               &     LSTM                & Word2Vec                                                     &             & \begin{tabular}[c]{@{}l@{}}Inv,FIBO,\\NYSE, BIS\end{tabular} \\ \hline
\end{longtable}
\end{landscape}
\clearpage
\twocolumn

\section{Dataset}
\label{sec:dataset}
In this section, we narrate the datasets we used to perform our experiments. In addition to the data, which was provided to us by the organizing team, we explored other external datasets as well. These include FIBO, DBpedia\cite{dbpedia}, Investopedia\footnote{\url{https://www.investopedia.com/}} and so on.

\subsection{Data Description}
\begin{figure}[ht]
  \centering
  \includegraphics[width=220pt]{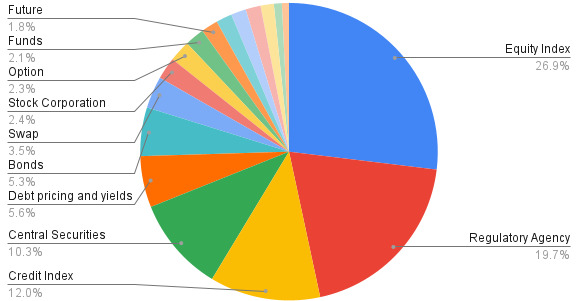}
  \caption{Distribution of labels in original training set}
  \label{fig:labeldist}
\end{figure}

\begin{figure*}[ht]
  \centering
  \includegraphics[width=\linewidth]{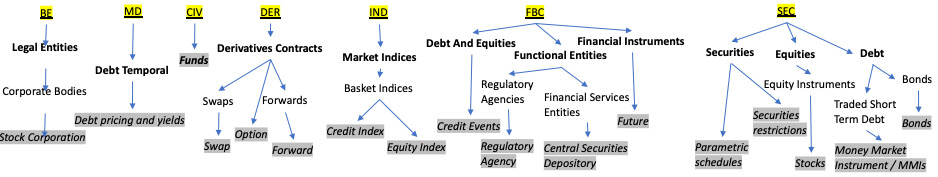}
  \caption{Hierarchy of labels as obtained from FIBO. Root nodes have been underlined and highlighted in yellow. First child nodes have been marked in bold. Leaf nodes have been italicised and highlighted in grey color.}
  \label{fig:fibo-label-hierarchy}
\end{figure*}

The organizers provided us with 211 prospectuses of different companies in Portable Document Format (PDF). Furthermore, a tagged dataset comprising 1050 financial terms and their corresponding hypernyms/labels were also provided. Out of 1050 terms, 1040 were distinct. We refer to this as the training set. Three of these terms were ambiguous as they were assigned 2 different labels. Terms with lengths less than or equal to six constituted more than 91\% of the training set. Number of distinct labels was 17. Their distribution is shown in Figure \ref{fig:labeldist} and presented in Table \ref{tab:labeldistri}. It is interesting to note that a hierarchy was present among these 17 labels as all of them belonged to FIBO. This hierarchy is presented in Figure \ref{fig:fibo-label-hierarchy}. The root nodes and leaf nodes have been highlighted in yellow and grey respectively. The first child nodes have been marked in bold. Moreover, we received 326 unlabeled financial terms which constituted the test set. 

\begin{table}
\centering
\caption{Distribution of labels in the original training set}
%\begin{tabular}{|l|l|}
\begin{tabular}{lr}
%\hline
\toprule
\textbf{Label}                & \textbf{Count} \\ %\hline
\midrule
Equity Index                  & 280            \\ %\hline
Regulatory Agency             & 205            \\ %\hline
Credit Index                  & 125            \\ %\hline
Central Securities Depository & 107            \\ %\hline %Central Securities Depository
Debt pricing and yields       & 58             \\ %\hline
Bonds                         & 55             \\ %\hline
Swap                          & 36             \\ %\hline
Stock Corporation             & 25             \\ %\hline
Option                        & 24             \\ %\hline
Funds                         & 22             \\ %\hline
Future                        & 19             \\ %\hline
Credit Events                 & 18             \\ %\hline
MMIs                          & 17             \\ %\hline
Stocks                        & 17             \\ %\hline
Parametric schedules          & 15             \\ %\hline
Forward                       & 9              \\ %\hline
Securities restrictions       & 8              \\ %\hline
\midrule
\textbf{Total}                & \textbf{1040}  \\ %\hline
\bottomrule
\end{tabular}
\label{tab:labeldistri}
\end{table}

\subsection{Data Augmentation} %[MORE CONTENT MAY BE ADDED IN FUTURE]}
Since 91\% of the financial terms in the training set had only six or fewer words, we explored various ways of augmenting the dataset. Similar approach was also followed by \cite{goat-2021-finsim} and \cite{saini-2020-anuj-investopedia}  while participating in FinSim-2 and FinSim-1 respectively. This was done in three phases. Let us understand each one of them.
\subsubsection{Acronym Expansion}
Several Financial Terms were present along with their acronyms. This led to inconsistency in the training set. Keswani et al. \cite{keswani-etal-2020-iitk} also highlighted this issue. To deal with this, we executed spaCy's\footnote{\url{https://spacy.io/}} inbuilt acronym detector on all the prospectuses. We manually investigated the outputs (i.e., a list of acronyms and their corresponding synonyms). We concluded that not all of outputs were usable. We developed the following heuristics to clean this list further. We dropped records having
\begin{itemize}
    \item expansions with number of characters lesser than that of the acronyms
    \item expansions with parenthesis/bracket symbols i.e., ``('' or ``)''
    \item expansions with number of characters lesser than or equal to five
    \item acronym which was a valid English word including proper nouns like ``bond'', ``England'', ``Germany'' and so on.
\end{itemize}
The cleaned list comprised 635 acronyms and their expansions. We used this list to augment our training set by replacing acronyms with their full forms wherever possible.
\subsubsection{Augmenting definitions from DBpedia}
\begin{figure*}[ht]
  \centering
  \includegraphics[width=\textwidth]{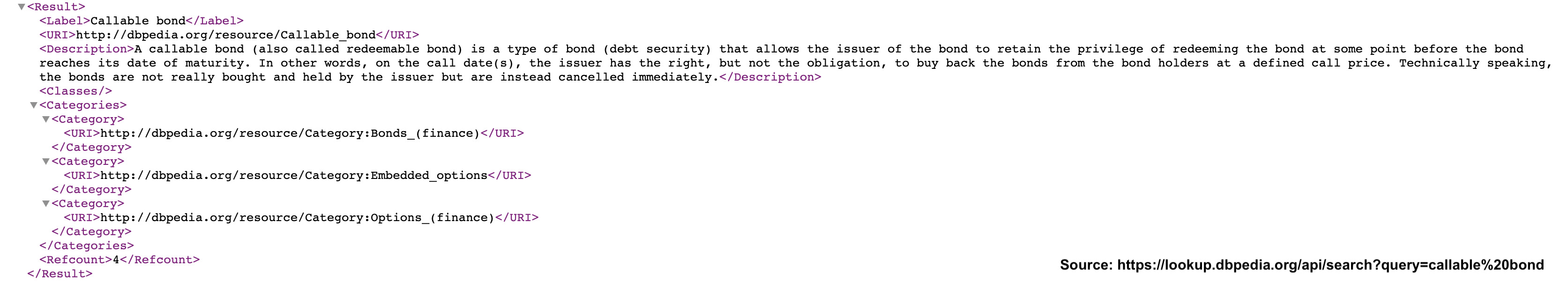}
  \caption{Result obtained by calling DBPedia Search API for the term ``callable bond''}
  \label{fig:dbpedia}
\end{figure*}

DBpedia\footnote{https://www.dbpedia.org/} provides search Application Programming Interfaces (API)\footnote{https://lookup.dbpedia.org/api/search} which helps in extracting structured information and relationships from Wikipedia\footnote{https://en.wikipedia.org/}. Kilger \cite{KLIEGR201559} introduced The Linked Hypernyms Dataset which provided more specific details than DBpedia.  We explored DBpedia extensively to obtain definitions of financial terms present in the training and test sets. These definitions added more context to the original terms. We present the results of invoking the search API for the term, ``callable bond'' in Figure \ref{fig:dbpedia}. Inspecting some of these sample outputs manually, we concluded that we needed to match the given financial terms with the content of the ``Label'' tag present in the output payloads and extract the contents of the ``Description'' tag. To achieve this, we pre-processed the given financial terms and the contents of the ``Label'' tag obtained by calling the search API for each of the terms. The pre-processing steps included conversion to lower case, punctuation and repetitive white space replacement and singularization. Furthermore, we calculated the token overlap ratio between these cleaned terms and contents of the ``Label'' tag using these formulas:\\
$Ratio1 = length(s_{1} \cap s_{2})/length(s_{1})$,\\
$Ratio2 = length(s_{2})/length(s_{1})$\\
where s\textsubscript{1} and s\textsubscript{2} represents sets of tokenized cleaned financial terms and tokenized cleaned contents of the ``Label'' tag respectively. After experimenting with several values, we empirically decided to use $Ratio1 = 1$ and $Ratio2 <= 1.25$. This enabled us to extract the descriptions of the matching terms from DBpedia.

\subsubsection{Augmenting definitions from Investopedia and FIBO}
While participating in FinSim-1, Saini \cite{saini-2020-anuj-investopedia} used definitions of financial terms from Investopedia\footnote{https://www.investopedia.com/financial-term-dictionary-4769738}. Inspired by his approach, we crawled all these definitions from Investopedia. A total of 6,261 definitions were obtained. Moreover, we obtained a glossary of 11,827 financial terms and their explanations from FIBO. We cleaned these using the approach mentioned previously. 

These data augmentation steps increased the size of the training set to 1836 records
and the size of the test set to 607 records. For the financial term ``callable bond'' we present the result of data augmentation in Table \ref{tab:sampledataaug}. Table \ref{tab:dataaug} presents the number of matches we get from different sources of data like DBPedia, Investopedia and so on.
 
\subsubsection{Adding data from various external sources}
\label{aug:other-sources}
Inspired by \cite{kroher-etal-2021-mxx}, we extracted 31,748 financial terms from various other websites such as 
\begin{itemize}
    \item Bank of International Settlements\footnote{\url{https://www.bis.org/regauth.htm}} (for label ``Regulatory Agency'')
    \item ETF Database\footnote{\url{https://etfdb.com/indexes/equity/}} (for label ``Equity Index'')
    \item Wikipedia\footnote{\url{https://en.m.wikipedia.org/wiki/Credit_default_swap_index}} \& Wiley\footnote{\url{https://onlinelibrary.wiley.com/doi/pdf/10.1002/9781119208631.app1}} (for label ``Credit Index'')
    \item Kaggle\footnote{\url{https://www.kaggle.com/stefanoleone992/mutual-funds-and-etfs/version/3?select=MutualFunds.csv}} (for label ``Funds'')
    \item ADVFN\footnote{\url{http://www.advfn.com/}} \& datahub\footnote{\url{https://datahub.io/core/nyse-other-listings}} (for label ``Stock Corporation'')
    \item National Securities Depository Limited\footnote{\url{https://nsdl.co.in/related/wrld.php}} and European Central Securities Depositories Association\footnote{\url{https://ecsda.eu/members-2/list-of-members}}  (for ``Central Securities'')
\end{itemize}
We added these terms to our training set for some of the experiments we performed. Later, we discarded them as it did not result in any improvement in the model performance. This is probably because most of these terms are proper nouns as they represent names of funds, organizations and so on.

\begin{table*}[ht]
\centering
\caption{Result obtained by data augmentation for the term ``callable bond''}
%\begin{tabular}{|l|l|l|}
\begin{tabular}{lll}
%\hline
\toprule
\textbf{Expanded Term/Term Definition} & \textbf{Label} & \textbf{Source}                                                           \\ %\hline
\midrule
Callable bond                          & Bonds          & \begin{tabular}[c]{@{}l@{}}original,\\ acronym\\ expansion\end{tabular} \\ \hline
\begin{tabular}[c]{@{}l@{}}Bond that includes a stipulation allowing the issuer \\ the right to repurchase and retire   the bond at the\\ call price after the call protection period\end{tabular} &
  Bonds &
  FIBO \\ \hline
\begin{tabular}[c]{@{}l@{}}A callable bond (also called redeemable bond) is a\\type of bond (debt security) that allows the issuer\\ of the bond to retain the privilege of redeeming\\ the bond at some  point before the bond reaches\\its date of maturity.\end{tabular} &
  Bonds &
  DBpedia \\ %\hline
\bottomrule
\end{tabular}
\label{tab:sampledataaug}
\end{table*}

\begin{table}
\centering
\caption{Number of matches obtained from various data sources}
%\begin{tabular}{|l|l|}
\begin{tabular}{lr}
%\hline
\toprule
\textbf{Data Source}    & \textbf{Count} \\ %\hline
\midrule
Original modelling data & 1040           \\ %\hline
Acronym expansion       & 218            \\ %\hline
DBpedia                 & 257            \\ %\hline
Investopedia            & 85             \\ %\hline
FIBO                    & 236            \\ %\hline
\bottomrule
\end{tabular}
\label{tab:dataaug}
\end{table}

\subsection{Development, Validation and Test splits}
As mentioned previously, we were provided with 1040 distinct manually tagged financial terms for training our model and 326 un-tagged instances for testing. We split the set of 1040 terms into two buckets: a development set having 831 terms (80\%) and a validation set having 209 terms (20\%). We did the same for the augmented set having 1836 financial terms out of which 1785 were distinct. This resulted in a set of 1440 distinct terms for training \& validation and a set of 345 distinct terms for testing. The final output i.e., predicted ranks of the given 17 labels on the test set was to be submitted for the initial set of 326 un-tagged instances. Thus, for the augmented test set we calculated the mean cosine similarity with each of the labels for multiple occurrences of a term. We ranked the labels based on these similarities.

The distribution of labels before (``original'') and after data augmentation (``extended'') is shown in Table \ref{tab:label-dis-before-after-aug}.

\begin{table*}
\centering
\caption{Label distribution for the development and validation set before and after data augmentation}
%\begin{tabular}{|l|l|l|l|l|}
\begin{tabular}{lrrrr}
%\hline
\toprule
                              & \multicolumn{2}{r}{\textbf{Original}} & \multicolumn{2}{r}{\textbf{Extended}} \\ %\hline
\midrule                              
\textbf{label}                & \textbf{\# dev}    & \textbf{\# val}   & \textbf{\# dev}    & \textbf{\# val}   \\ %\hline
\midrule
Equity Index                  & 225                & 57                & 373                & 84                \\ %\hline
Regulatory Agency             & 159                & 46                & 260                & 78                \\ %\hline
Credit Index                  & 103                & 21                & 123                & 27                \\ %\hline
Central Securities Depository & 83                 & 24                & 106                & 28                \\ %\hline
Bonds                         & 49                 & 6                 & 110                & 14                \\ %\hline
Debt pricing and yields       & 41                 & 17                & 84                 & 34                \\ %\hline
Swap                          & 31                 & 5                 & 57                 & 9                 \\ %\hline
Option                        & 21                 & 3                 & 35                 & 4                 \\ %\hline
Stock Corporation             & 18                 & 6                 & 54                 & 15                \\ %\hline
Funds                         & 17                 & 5                 & 36                 & 10                \\ %\hline
Future                        & 16                 & 3                 & 29                 & 7                 \\ %\hline
Credit Events                 & 15                 & 3                 & 35                 & 6                 \\ %\hline
Parametric schedules          & 14                 & 1                 & 45                 & 3                 \\ %\hline
MMIs                          & 14                 & 3                 & 29                 & 9                 \\ %\hline
Stocks                        & 12                 & 5                 & 23                 & 11                \\ %\hline
Securities restrictions       & 7                  & 1                 & 28                 & 3                 \\ %\hline
Forward                       & 6                  & 3                 & 13                 & 3                 \\ %\hline
\midrule
\textbf{TOTAL}                & \textbf{831}       & \textbf{209}      & \textbf{1440}      & \textbf{345}      \\ %\hline
\bottomrule
\end{tabular}
\label{tab:label-dis-before-after-aug}
\end{table*}

\section{Methodology}
\label{sec:meth}
\begin{figure*}[ht]
  \centering
  \includegraphics[width=\textwidth]{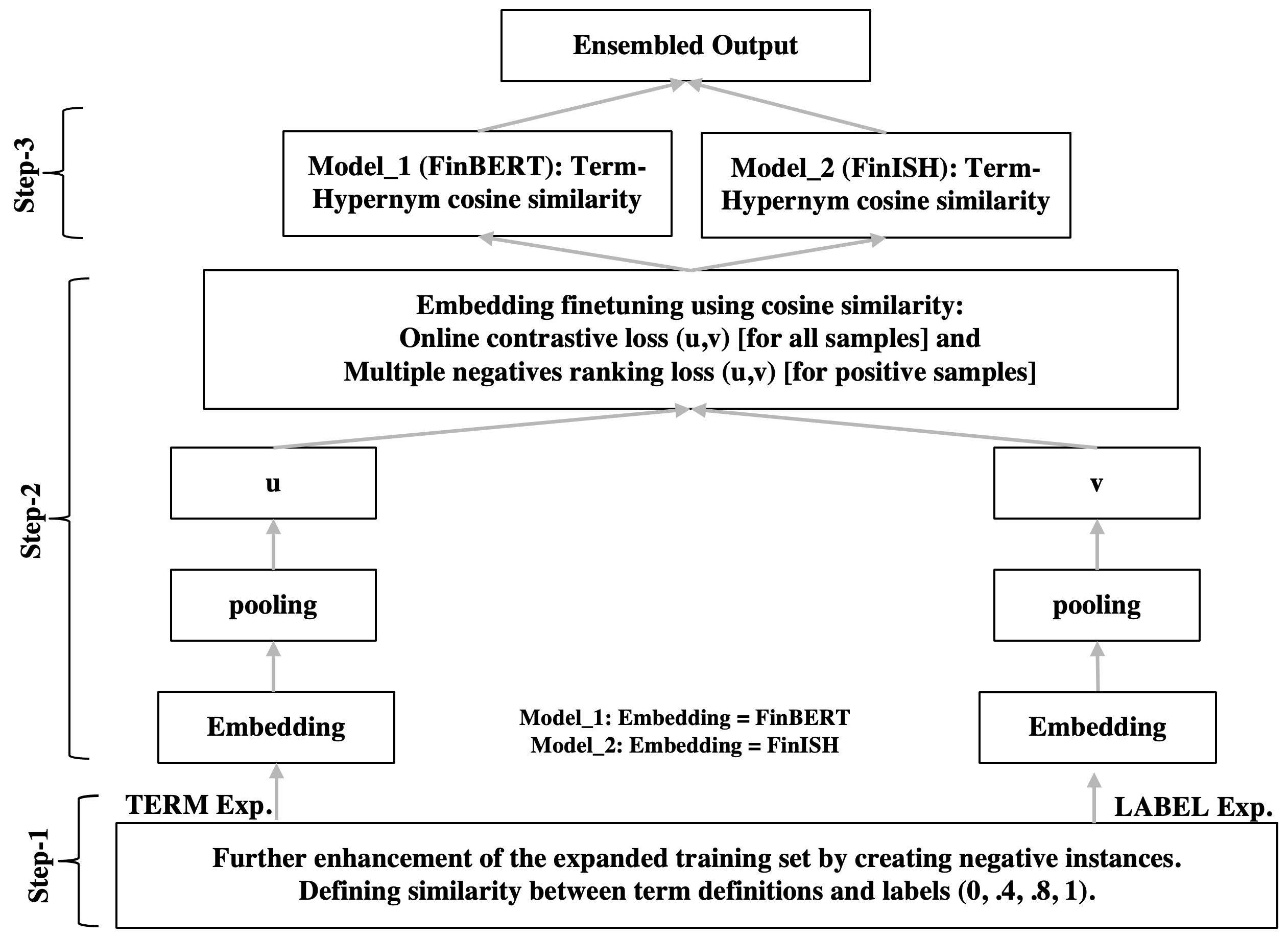}
  \caption{Methodology}
  \label{fig:finsim-meth}
\end{figure*}

Our best performing model is an ensemble of two models. Each of these models has been developed in three steps.
\begin{enumerate}
    \item negative sample creation
    \item using sentence transformers to fine-tune embeddings having 768 dimensions (reference: Algorithm \ref{alg:neg_set})
    \item calculating cosine similarities between terms and hypernyms.
\end{enumerate}
This has been depicted in Figure \ref{fig:finsim-meth}. Steps 1 and 3 are common for both models. In the second step, we use FinBERT \cite{araci2019finbert} embeddings for the first model and FinISH \cite{akl-etal-2021-yseop} embeddings for the  second model.

\textbf{\underline{STEP-1}}: In the first step, we create negative samples from the existing training set having sets of terms `T', labels `L', term definitions `TT' and label definitions `LL'. The definitions of labels and terms are obtained through data augmentation. For instances where we are not able to augment anything to a given financial term, we keep the term definition the same as the term. For each term `t' having definition `td', its corresponding label `l' and label definition `ld', present in the training set we first assign a similarity score of 1.0  to the (`td', `ld') pair. After that, we extract root node `ln' and first child node `lc' of `l'. We then randomly select 10 labels and their corresponding definitions from `L' such that none of the selected labels and their corresponding terms is the same as `l' and `t'. For each such label `la' and label definition `lnd', we assign similarity scores corresponding to each of the (`td', `lnd') pairs. This similarity score is assigned a value based on the following conditions\\ i) value = 2.0*k when the first child of `la' i.e. `lac' is the same as `lc'\\ ii) value = 1.0*k when only the root node of `la' i.e. `lan' is same as `ln' and its first child `lac' is different from `lc'\\ iii) value = 0.0*k when former two conditions are not met i.e. they have no ancestors in common\\ We present this formally in Algorithm \ref{alg:neg_set}. We empirically determine that keeping the value of parameter k as 0.4 gives the best result. This resulted in 63,360 instances in total out of which 49,836 had a similarity score of 0.0. We sub-sampled the instances with similarity score of 0.0. The final distribution consists of 5,760 instances with a 1.0  similarity score, 5304 instances with 0.8, 2460 with 0.4 and 550 with a similarity score of 0.0. This step is common for both the models described above.

\textbf{\underline{STEP-2}}: In the second step, for the first model we fine-tune FinBERT \cite{araci2019finbert} embeddings using sentence transformer \cite{sbert} architecture. For the second model, we further fine-tune the FinISH embeddings released by Yseop Labs\cite{akl-etal-2021-yseop}. They created this embedding by fine-tuning RoBERTa\cite{Liu2019RoBERTaAR} on the FIBO corpus. Our objective was to minimize the multiple negative ranking loss and online contrastive loss. We kept the margin parameter at 0.5. A batch size of 20, when executed for 25 epochs, gave the best result for the first model. For the second model, a batch of 30 when executed for 45 epochs gave the best result. The sample code is available here.\footnote{\url{https://www.sbert.net/examples/training/quora_duplicate_questions/README.html##multi-task-learning}\\ (accessed on October 2021)}.

\textbf{ \underline{STEP-3}}: In the third step, we convert definitions of all the 17 labels/hypernyms and terms present in the validation and test set into vectors. We use the fine-tuned embeddings generated in the previous step for the same. We further calculate cosine similarity between the vectors of each of these terms with that of all the 17 hypernyms. Since we have had augmented the dataset, we need to roll up this data such that we have only one record for every term. We use the mean of cosine similarities to achieve this. We do the same for the other model as well. This results in two cosine similarities for each of the terms one obtained from the first model while the other from the second. 

To ensemble, we again take the mean of the two cosine similarities we calculated for each of the terms across all the hypernyms. Finally, we rank the hypernyms in terms of decreasing order of the mean cosine similarity.

\begin{algorithm*}
\caption{Algorithm to generate negative samples from existing training set}\label{alg:neg_set}
\begin{algorithmic}[1]
\Require $T > 0$ and $L > 0$ \Comment{T is the augmented set of financial terms and L consists of corresponding labels i.e., hypernyms. $TT > 0$ and $LL > 0$ are the set of definitions of terms and labels respectively obtained after performing data augmentation}
\Require Function FR(n) and Function FC(n) \Comment{Function FR and FC returns the root node and first child node corresponding to node n respectively where n is one of the 17 labels i.e., leaf nodes/hypernyms}
\Ensure $length(T) = length(TT) = length(L) = length(LL)$
\State $NT \gets \{\}$  \Comment{NT is the new set of definitions of financial terms to be created by appending negative samples}
\State $NL \gets \{\}$ \Comment{NL is the new set of definitions of labels corresponding to terms in NT}
\State $NS \gets \{\}$ \Comment{NS is the set of assigned similarity scores between the newly selected definitions of terms and labels in NT \& NL respectively}
\State $k \gets 0.0$ \Comment{`k' is a hyper-parameter. Keeping k = 0.0 gives the best result}
\For{each term t $\in$ T, term definition td $\in$ TT, corresponding label l $\in$ L and label definition ld $\in$ LL}
    \State $NT \gets NT \cup \{td\}$
    \State $NL \gets NL \cup \{ld\}$
    \State $NS \gets NS \cup \{1.0\}$ \Comment{Assign a similarity score of 1.0 as the term and the label definition belong to the original set}
    \State $ln \gets FR(l)$ \Comment{Extract root node of label `l'}
    \State $lc \gets FC(l)$ \Comment{Extract first child node of label `l'}
    %\State $cnt \gets 1$
    \State{R, RR $\varepsilon_\mathrm{r}$ L, LL where length(R)=10, length(RR)=10 } \Comment{Randomly select 10 labels from `L 'and corresponding label definitions from `LL' ensuring none of the labels are `l' and none of their corresponding terms is `t'. This is done for creating the negative set}
    \For{each label la $\in$ R and corresponding definition lnd $\in$ RR}
    %\While{$cnt \leq 10$}
    %\State{la $\varepsilon_\mathrm{r}$ L} \Comment{Randomly select label `la' from L to create negative set}
            %\State $cnt \gets cnt + 1$
            %\If{$la \neq l$}
                \State $NT \gets NT \cup \{td\}$
                \State $NL \gets NL \cup \{lnd\}$
                \State $lan \gets FR(la)$ \Comment{Extract root node of label `la'}
                \State $lac \gets FC(la)$ \Comment{Extract first child node of label `la'}
                \If{$lac = lc$} \Comment{Check if first child nodes are the same. This implies root nodes are also the same.}
                    \State $NS \gets NS \cup \{2 * k\}$
                \ElsIf{$lan = ln$} \Comment{Check if root child nodes are same when first child nodes are different}
                    \State $NS \gets NS \cup \{1 * k\}$
                \Else \Comment{When first child nodes and root nodes are different}
                    \State $NS \gets NS \cup \{0 * k\}$
                \EndIf
            %\Else
            %    \State continue;
            %\EndIf
    %\EndWhile
    \EndFor
\EndFor
\State \Return $NT, NL, NS$
\end{algorithmic}
\end{algorithm*}

\section{Experimentation}
\label{sec:experimentation}
In this section, we shall narrate various experiments we performed systematically to arrive at final model described in the previous section. We started by evaluating the baseline models provided to us.

\subsection{Baselines}
Let's understand the baseline solutions provided by the organizers. Kang et al. \cite{kang-etal-2021-finsim} trained a custom word2vec \cite{word2vec} model having 300 dimensions on text corpus extracted from the prospectus. 
\subsubsection{Baseline-1} In the first system, they calculate distances between terms and hypernyms based on the custom word2vec embeddings. They rank the hypernyms on the increasing order of distance.

\subsubsection{Baseline-2} The second system comprises a logistic regression-based classifier trained using custom word2vec embeddings of the financial terms as independent variables and hypernyms as the dependent variables.

\subsection{Experiments}
At first, we removed the duplicate observations that we observed in the original dataset. We reserved 20\% of the data for the unbiased validation set and the remaining 80\% was used for training the models. We identified sources like DBpedia, FIBO and Investopedia which contain the definitions of many terms present in the input set. We also extracted the acronym definitions from the prospectus corpus shared by the organizers. All these sources helped us to augment the training data. The augmented data consisted of the original records along with the records where input terms were replaced with definitions and expansions. The number of instances in the original and the augmented training set was 832 and 1470. Similarly, the number of instances in the original and the augmented validation set was 208  and 366. This indicates we were not able to get a definition or expansion for each of the terms.

We began the experimentation by creating Term Frequency Inverse Document Frequency (TF-IDF) matrix, Topic Models and creating a machine learning based classifier over it. Since the performance was not appealing, we fine-tuned one of the State of The Art (SOTA) pre-trained models known as BERT \cite{devlin-etal-2019-bert}. We used sub-word tokenization and followed the standard classification architecture to fine-tune the pre-trained models. We took the representation from [CLS] token and passed it to the feedforward layers. The last layer of the network had 17 nodes with SoftMax activation. These 17 nodes provided the prediction for 17 labels mentioned previously. We did not freeze the base model while training. This enabled the fine-tuning of the base model for the task at hand, resulting in better performance. During the training, the error was propagated back through the transformer network. Looking at the distribution of the tokenized output length, we decided to keep the maximum input sequence length as 32. We ran extensive hyperparameter tuning and identified that a combination of Adam optimizer with a learning rate of 0.00002 and 64 batch size gave us the best result. We trained the model for 40 epochs and based on the validation set performance. It performed the best after the 18\textsuperscript{th} epoch. We ordered the hypernyms in decreasing order of predicted probabilities. This performance was much better than that of the baselines. 

We further tried the same BERT model in the augmented dataset which included the definitions from various sources mentioned previously. These definitions were well-structured sentences and they comprised longer sequences of input terms. We repeated the experiments described previously after increasing the input maximum input sequence length to 256. This input length was decided based on the distribution of the number of tokens that were present in the term definitions after the augmentation step. We trained it till the 40 epoch and found out that its performance on the validation set was best at the 17\textsuperscript{th} epoch. We observed that this performance was significantly better than that of the models developed without data augmentation. This led us to conclude that the data augmentation steps we followed were useful. We also tried adding data from various other sources as mentioned in section \ref{aug:other-sources}. However, this did not yield any further improvement in the performance of the model. This is probably because most of these terms were proper nouns and organization like entities. 

We subsequently tried out various other transformer-based models present in the Huggingface \cite{wolf-etal-2020-transformers} model repository. This included RoBERTa \cite{Liu2019RoBERTaAR}, FinBERT \cite{araci2019finbert}, FinEAS \cite{gutierrezfandino2021fineas} and so on. We observed that FinBERT when fine-tuned using the expanded data set further improved the performance. Subsequently, we trained a new model based on transformer architecture. Its objective was to predict two things together i) root node ii) hypernyms. This did not perform well. We also tried to fine-tune these models using the Masked Language Model based approach on the corpus of the prospectus. Due to resource constraints,  we could not train it beyond a few epochs. Its performance was not promising as well.

After extensively studying the failed cases and observing the hierarchy of the labels we decided to try out a novel framework to generate negative instances and fine-tune it, using the sentence transformer \cite{sbert} architecture. This has been elaborated in detail in section \ref{sec:meth}. For creating the negative set mentioned in Algo: \ref{alg:neg_set}, we experimented with different sampling strategies and with various values of `k'. The performance of the model improved when we used the sentence transformer architecture with FinBERT at the back end. It improved further on changing the base embedding from FinBERT to FinISH. FinISH was developed and resealed by Yseop Labs\footnote{\url{https://yseop.com/}} while participating in FinSim-3 \cite{akl-etal-2021-yseop}. We ran it for 45 epochs with a batch size of 30. It took around 1 hour 43 minutes to train.

Finally, we tried to ensemble the best performing models. We observed that an ensemble of the last two models which were trained using sentence transformers architecture with negative samples resulted in the best performance on the validation set. All the hyper-parameters were selected empirically by tracking the model performance on the validation set.

\subsection{Implementation Details}
%GPU, Python, Colab, Other methods tried
We performed the experiments on Google Colab\footnote{\url{https://research.google.com/colaboratory/}} (free tier) and on a Nvidia DGX GPU cluster. The cluster consists of 32 Nvidia Tesla V100 GPUs, over 160,000 CUDA
cores and over 20,000 Tensor Cores. We used Python (3.7) for all the computations. The main libraries used here consists of PyTorch\footnote{\url{https://pytorch.org/}}, SentenceTransformers\footnote{\url{https://www.sbert.net/}}, pandas\footnote{\url{https://pandas.pydata.org/}}, NumPy\footnote{\url{https://numpy.org/}} and scikit-learn\footnote{\url{https://scikit-learn.org/stable/}}.

\subsection{Challenges}
Just like most other data science problems, this problem consisted of two major challenges. They are limited data and computational resources.  The set of financial terms provided to us had few tokens. We have overcome this by augmenting data from various other sources as discussed earlier. We moved from Google Colab (free tier) to the GPU cluster mentioned above to deal with the computational limitations.

\section{Results and Discussions}
\label{sec:results}
In this section, we shall discuss the results presented in Table \ref{tab:results}. %\textcolor{blue}{
We restrict our evaluation to just one dataset due to non-availability of any other dataset suitable for financial hypernym detection.%}
Models with serial numbers (SLN) 1 to 15 were developed during the FinSim-3 challenge while those with SLN 16 to 20 were developed later. After the event, the organizers declared the results for each submission of the participating teams. The number of submissions was restricted to 3.  Thus, we present test set results for three of our models (SLN: 5, 6, 7). On comparing this with the test set results of other participants (SLN: 8 to 15), we observe that 
%%%our old model SFinBERT\_neg (SLN: 7) \cite{chopra-ghosh-2021-term} %% Change SOHOM
our old model SFinBERT\_neg (SLN: 7) \cite{chopra-ghosh-2021-term}
ranked third and was marginally behind the one which was ranked second (SLN: 15) \cite{akl-etal-2021-yseop}. This model was developed by fine-tuning FinBERT \cite{araci2019finbert} with negative samples using sentence transformer architecture. We tried reaching the organisers to evaluate our new model (SLN: 20) on the test set as well. However, the test set has not yet been released publicly. Thus, we present our results on the holdout validation set.

It is interesting to observe that on using transformer-based pre-trained BERT embeddings (SLN: 3, 4), the model performs better than the baselines (SLN: 1, 2). This proves the effectiveness of transformer-based embeddings like BERT \cite{devlin-etal-2019-bert} over traditional embeddings like word2vec \cite{word2vec}. It happened probably because transformer-based embeddings having been pre-trained on large datasets can capture more complexities within the language. Comparing the performance of models (having SLN: 3 and 5) with those (having SLN: 4 and 6) we conclude that external data augmentation has resulted in a performance gain. We also notice that financial domain specific embedding FinBERT \cite{araci2019finbert} (SLN: 5, 6) resulted in improvement of the model performance when compared to generic embedding like BERT \cite{devlin-etal-2019-bert} (SLN: 3, 4). Furthermore, it is quite interesting to note that fine-tuning FinBERT \cite{araci2019finbert} using a classifier layer to top (SLN: 5 and 6) to predict hypernym did not perform as good as fine-tuning a FinBERT model using sentence transformer where negative samples were also included (reference: SFinBERT\_neg with SLN: 7). This is because several hypernyms were inter-dependent as shown in Figure \ref{fig:fibo-label-hierarchy}.

Models with SLN 8 to 15 have been developed by other participating teams. Since their models were not open sourced, we are not able to present the performance of their models on our hold-out validation set. %\textcolor{blue}{
For the team MXX (SLN: 13), we quote the performance on their validation set as presented in the paper \cite{kroher-etal-2021-mxx}.%}.
We mentioned the approaches followed by other teams in Table \ref{tab:related-works-finsim}. In the model SFinBERT\_neg\_th (SLN: 16) we changed `k' (mentioned in section \ref{sec:meth}) from 0.4 to 0.2. The rest has been kept the same as the model SFinBERT (SLN: 7). Similarly, we tried changing the sampling strategy in the model SFinBERT\_neg\_ss (SLN: 17). Instead of sampling over the entire set `L' (as mentioned in Algorithm \ref{alg:neg_set}), we tried considering all other hypernyms. Both methods did not improve the performance.

Moreover, in the model SFinBERT\_neg (SLN: 7) we tried using FinISH embeddings instead of the FinBERT embeddings. We trained it for 45 epochs after increasing batch size to 30. This improved the model performance (Mean Rank: 1.072 and Accuracy: 0.952). We refer this model as SFinHyp\_neg (SLN: 18). As mentioned in section \ref{aug:other-sources}, on adding more data to this model deteriorated the performance slightly. This is due to the fact this data is comprised mainly of proper nouns. We refer to it as Model SFinHyp\_more\_data (SLN: 19).
Finally, ensembling models SFinHyp\_neg (SLN: 18) with SFinBERT\_neg (SLN: 7) resulted in the best performance (Mean Rank: 1.053 and Accuracy: 0.967). It performed even better than the old model we submmited at FinSim-3 (SLN: 7) %\textcolor{blue}{
and the existing SOTA model MXX (SLN: 13)%}
on the held out validation set. We denote this ensemble model as Ensemble\_7\_18 (SLN: 20).

We further analyse the results for every label along with their root nodes. This is presented in Table \ref{tab:results-labelwise}. We notice that for all the labels having root node `CIV', `SEC' and for labels `Forward', `Option', `Future', `Credit Events' and `Equity Index' the model performs the best. For the labels `Stock Corporation', `Swap' the proposed model performs the worst. For all other labels, the model performance is mediocre. 

As a next step, we used Principal Component Analysis (PCA) to visualize the embeddings of the hypernyms generated using the method SFinHyp\_neg (SLN: 18) in 2 dimensions. It is quite interesting to note that `Option' and `Future' despite having neither the root node nor the first child node in common are close to each other. This is because they are similar financial trading products. Thus, we can say the model captured the semantic aspect to some extent as well. We also observe that `Regulatory Agency' and `Central Securities Depository' which have the same root node `FBC' are together. Similarly, hypernyms which do not have anything in common like `Stock Corporation' and `Debt pricing and yields' are separate from the rest. However, this is not the case for most other hypernyms. This is because we are losing out on much information while projecting 768 dimensions of the embeddings to 2 dimensions. Our PCA model captures only 28.3\% of the variance.

\begin{figure*}[ht]
  \centering
  \includegraphics[width=\textwidth]{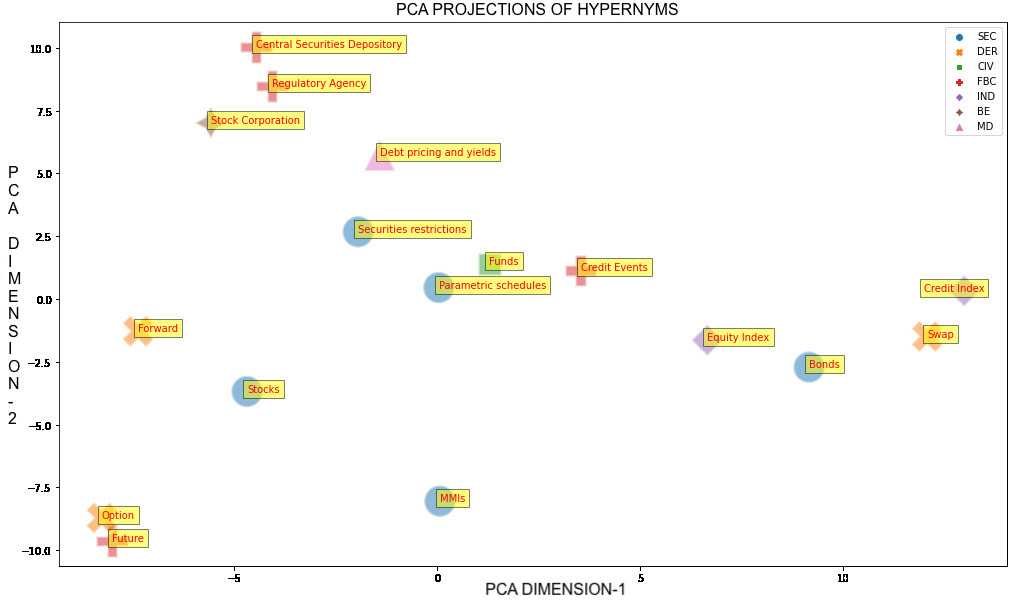}
  \caption{PCA projection of embeddings of Hypernyms in 2 dimensions. Same shape denotes same root nodes.}
  \label{fig:finsim-tsne}
\end{figure*}

\begin{table*}
\centering
\caption{Results on validation and test set. Org. represents original and Ext. represents extended. Base refers to baseline. MR is Mean Rank}
\begin{tabular}{llrrrrr}
%\hline
\toprule
   &                      &     & \multicolumn{2}{r}{\textbf{Validation Set}} & \multicolumn{2}{r}{\textbf{Test Set}}               \\ %\hline
   \midrule
\textbf{SLN.} &
  \textbf{Model} &
  \textbf{Data Aug.} &
  \multicolumn{1}{r}{\textbf{MR}} &
  \textbf{Acc.} &
  \multicolumn{1}{r}{\textbf{MR}} &
  \textbf{Acc.} \\ %\hline
  \midrule
1  & Base-1               & No  & \multicolumn{1}{r}{2.158}      & 0.498      & \multicolumn{1}{r}{1.941}          & 0.564          \\ %\hline
2  & Base-2               & No  & \multicolumn{1}{r}{1.201}      & 0.876      & \multicolumn{1}{r}{1.750}           & 0.669          \\ %\hline
3  & BERT                 & No  & \multicolumn{1}{r}{1.177}      & 0.899      & \multicolumn{1}{r}{-}              & -              \\ %\hline
4  & BERT                 & Yes & \multicolumn{1}{r}{1.153}      & 0.928      & \multicolumn{1}{r}{-}              & -              \\ %\hline
5  & FinBERT              & No  & \multicolumn{1}{r}{1.117}      & 0.928      & \multicolumn{1}{r}{1.257}          & 0.886          \\ %\hline
6  & FinBERT              & Yes & \multicolumn{1}{r}{1.110}       & 0.942      & \multicolumn{1}{r}{1.220}           & 0.895          \\ %\hline
7  & SFinBERT\_neg (Our old model) \cite{chopra-ghosh-2021-term}       & Yes & \multicolumn{1}{r}{1.086}      & 0.947      & \multicolumn{1}{r}{1.156}          & 0.917          \\ %\hline
8  & dicoe\_1  \cite{loukas-etal-2021-dicoe}           & No  & \multicolumn{1}{r}{-}          & -          & \multicolumn{1}{r}{1.180}           & 0.889          \\ %\hline
9  & dicoe\_2  \cite{loukas-etal-2021-dicoe}           & Yes & \multicolumn{1}{r}{-}          & -          & \multicolumn{1}{r}{1.162}          & 0.904          \\ %\hline
10 & MiniTrue\_2 \cite{feng-wei-2021-exploiting}         & No  & \multicolumn{1}{r}{-}          & -          & \multicolumn{1}{r}{1.315}          & 0.865          \\ %\hline
11 & MiniTrue\_1 \cite{feng-wei-2021-exploiting}        & No  & \multicolumn{1}{r}{-}          & -          & \multicolumn{1}{r}{1.346}          & 0.855          \\ %\hline
12 & MiniTrue\_3 \cite{feng-wei-2021-exploiting}         & No  & \multicolumn{1}{r}{-}          & -          & \multicolumn{1}{r}{1.337}          & 0.825          \\ %\hline
13 & mxx \cite{kroher-etal-2021-mxx}                 & Yes & \multicolumn{1}{r}{\textit{1.06}}          & \textit{0.96}          & \multicolumn{1}{r}{\textbf{1.113}} & \textbf{0.941} \\ %\hline
14 & yseop\_1 \cite{akl-etal-2021-yseop}             & Yes & \multicolumn{1}{r}{-}          & -          & \multicolumn{1}{r}{1.236}          & 0.883          \\ %\hline
15 & yseop\_2 \cite{akl-etal-2021-yseop}             & Yes & \multicolumn{1}{r}{-}          & -          & \multicolumn{1}{r}{1.141}          & 0.917          \\ %\hline
16 & SFinBERT\_neg\_th    & Yes & \multicolumn{1}{r}{1.110}       & 0.938      & \multicolumn{1}{r}{-}              & -              \\ %\hline
17 & SFinBERT\_neg\_ss    & Yes & \multicolumn{1}{r}{1.105}      & 0.933      & \multicolumn{1}{r}{-}              & -              \\ %\hline
18 & SFinHyp\_neg & Yes & \multicolumn{1}{r}{1.072}      & 0.952      & \multicolumn{1}{r}{-}              & -              \\ %\hline
19 & SFinHyp\_more\_data  & Yes & \multicolumn{1}{r}{1.306}      & 0.813      & \multicolumn{1}{r}{-}              & -              \\ %\hline
20 &
  \textbf{Ensemble\_7\_18 (Our new Model)} &
  Yes &
  \multicolumn{1}{r}{\textbf{1.053}} &
  \textbf{0.967} &
  \multicolumn{1}{r}{-} &
  - \\ %\hline
  \bottomrule
\end{tabular}
\label{tab:results}
\end{table*}

\begin{table}
\centering
\caption{Model performance for each labels\\ CSD means Central Securities Depository}
\label{tab:results-labelwise}
%\begin{tabular}{|l|l|l|l|}
\begin{tabular}{llll}
%\hline
\toprule
\textbf{Root} & \textbf{Label}                & \textbf{Mean Rank} & \textbf{Acc.} \\ %\hline
\midrule
BE            & Stock Corporation             & 1.333              & 0.833         \\ %\hline
CIV           & Funds                         & 1.000              & 1.000         \\ %\hline
DER           & Forward                       & 1.000              & 1.000         \\ %\hline
DER           & Option                        & 1.000              & 1.000         \\ %\hline
DER           & Swap                          & 1.200              & 0.800         \\ %\hline
FBC           & Future                        & 1.000              & 1.000         \\ %\hline
FBC           & Regulatory Agency             & 1.087              & 0.935         \\ %\hline
FBC           & CSD & 1.042              & 0.958         \\ %\hline
FBC           & Credit Events                 & 1.000              & 1.000         \\ %\hline
IND           & Equity Index                  & 1.000              & 1.000         \\ %\hline
IND           & Credit Index                  & 1.143              & 0.952         \\ %\hline
MD            & Debt pricing and yields       & 1.059              & 0.941         \\ %\hline
SEC           & Bonds                         & 1.000              & 1.000         \\ %\hline
SEC           & MMIs                          & 1.000              & 1.000         \\ %\hline
SEC           & Stocks                        & 1.000              & 1.000         \\ %\hline
SEC           & Parametric schedules          & 1.000              & 1.000         \\ %\hline
SEC           & Securities restrictions       & 1.000              & 1.000         \\ %\hline
\bottomrule
\end{tabular}
\end{table}

\subsection*{Ablation Study}
To understand the significance of each component of our model (Ref: Figure \ref{fig:finsim-meth}) we do an ablation study. We present the results in Table \ref{tab:ablation-study}. Analysing these results, we see that if we use readily available FinBERT embeddings \cite{araci2019finbert} or fine-tuned RoBERTa embeddings \cite{akl-etal-2021-yseop} to simply rank the hypernyms based on cosine similarity with the financial terms and their definitions, then the performance deteriorates drastically. This explains the importance of the algorithm we developed to create negative sets. The final ensemble model performs better than the constituent models.

\begin{table}
\centering
\caption{Ablation Study on the validation set. \\cos. sim. means cosine similarity}
%\begin{tabular}{|l|ll|}
\begin{tabular}{lrr}
%\hline
\toprule
%\textbf{}                     & \multicolumn{2}{c|}{\textbf{Validation Set}}             \\ \hline
\textbf{Model}                & \textbf{Mean Rank} & \textbf{Acc.}  \\ %\hline
\midrule
Only FinBERT + cos. sim.  & 2.421              & 0.297          \\ %\hline
Only SFinHyp + cos. sim.  & 1.301              & 0.804          \\ %\hline
SFinBERT\_neg                 & 1.086              & 0.947          \\ %\hline
SFinHyp\_neg                  & 1.072              & 0.952          \\ %\hline
\textbf{Ensemble (Our Model)} & \textbf{1.053}     & \textbf{0.967} \\ %\hline
\bottomrule
\end{tabular}
\label{tab:ablation-study}
\end{table}

\section{Conclusion}
\label{sec:conclusion}
In this paper, we study the approaches followed by participants of all three editions of the FinSim challenge. Furthermore, we present a novel method of fine-tuning FinBERT \cite{araci2019finbert} and FinISH \cite{akl-etal-2021-yseop} embeddings using hierarchies present in FIBO. This enabled us to rank a set of hypernyms for a given financial term. We conclude that pre-trained transformer-based embeddings fine-tuned with domain specific data performed better in this scenario. We also observe that augmenting the existing data set with external data enhanced the model performance. However, adding more data like names of companies, mutual funds and stocks did not add any value.

While studying the stability of the model, we observe that during the training phase, we picked up random samples only in two places. During evaluation, we use two models to generate embeddings. These are further used to calculate cosine similarities between a given set of financial terms and hypernyms. The final ranking is done by taking mean of these two cosine similarities. Thus, the predictions generated from the ensemble model are stable.

Unlike the models developed by other participating teams (\cite{kroher-etal-2021-mxx}, \cite{akl-etal-2021-yseop} and so on), our model is not a classification model. Thus, we don't need to retrain it frequently if additional hypernyms are added. Moreover, the LSTM network which team MXX \cite{kroher-etal-2021-mxx} trained cannot be parallelized and scaled. It won't be able to effectively deal with out-of-vocabulary words. It is easier to compute the mean of two cosine similarities than using two bi-directional LSTM networks to predict the hypernyms. This makes our model simple, scalable and easy to deploy when compared to that of the others.

\section{Future Works}
\label{sec:future}
In future, we would like to gather more data for training and explore the use of Knowledge Graphs and Graph Neural Networks to improve these models. We also want to work on interpreting these models using various model explainability plots and participate in the upcoming challenges like FinSim-4\footnote{\url{https://sites.google.com/nlg.csie.ntu.edu.tw/finnlp-2022/shared-task-finsim4-esg}}. Furthermore, an interesting direction for further research would be to create embeddings especially for financial terms and their definitions. Presently, we explored the hierarchies and relation trees present in FIBO. Although `Future' and `Options' are similar trading products, they are present in different trees. We would like to take this into account as well while creating our negative set.  Using Neural Network based ranking loss may result in the better rank ordering of the hypernyms. Finally, we want to evaluate the statistical significance of predictions from these models over the baselines on a larger dataset.

\section*{Conflict of interest} On behalf of all authors, the corresponding author states that there is no conflict of interest.

\section*{Declarations}
\subsection*{Ethics approval} This research did not involve any human participants and/or animals. There was no need for informed consent.
\subsection*{Funding} Not Applicable.
\subsection*{Author contributions} Sohom Ghosh and Ankush Chopra conducted the experiments and prepared the manuscript. Sudip Kumar Naskar re-examined it. All authors reviewed the manuscript.
\subsection*{Availability of data and material} The data set used in this paper can be obtained from \url{https://sites.google.com/nlg.csie.ntu.edu.tw/finnlp2021/shared-task-finsim}. The metadata is presented in the paper \cite{kang-etal-2021-finsim}.
\subsection*{Code availability} Our code base will be made available in \url{https://github.com/sohomghosh/FinSim_Financial_Hypernym_detection} after acceptance of the paper.
\subsection*{Acknowledgements}
%%We acknowledge Fidelity Investments for providing us with resources for computation. %% MAY REMOVE FIDELITY %%SOHOM
We express our sincere gratitude to the organizers of FinSim-3 \cite{kang-etal-2021-finsim} for providing us with labelled data, evaluation scripts and starter codes. 
%We thank the anonymous reviewers for providing us with valuable feedbacks.

\bibliography{sn-bibliography}


\end{document}